\documentclass[acmtog, nonacm]{acmart}
\makeatletter
\let\@authorsaddresses\@empty
\makeatother

\usepackage{booktabs} 
\usepackage{float}
\usepackage{multirow}

\newcommand\scalemath[2]{\scalebox{#1}{\mbox{\ensuremath{\displaystyle #2}}}}

\DeclareMathOperator{\Sim}{Sim}
\DeclareMathOperator{\SO}{SO}

\usepackage[ruled]{algorithm2e} 

\SetAlFnt{\small}
\SetAlCapFnt{\small}
\SetAlCapNameFnt{\small}
\SetAlCapHSkip{0pt}

\usepackage{comment}

\usepackage{xcolor} 
\usepackage{hyperref}
\let\oldcite\cite
\renewcommand{\cite}[1]{\textcolor{cyan}{\oldcite{#1}}}

\begin{document}

\title{BAG: Body-Aligned 3D Wearable Asset Generation}

\begin{teaserfigure}
  \centering
  \hspace*{-0.8cm}
  \includegraphics[width=1.08\linewidth]{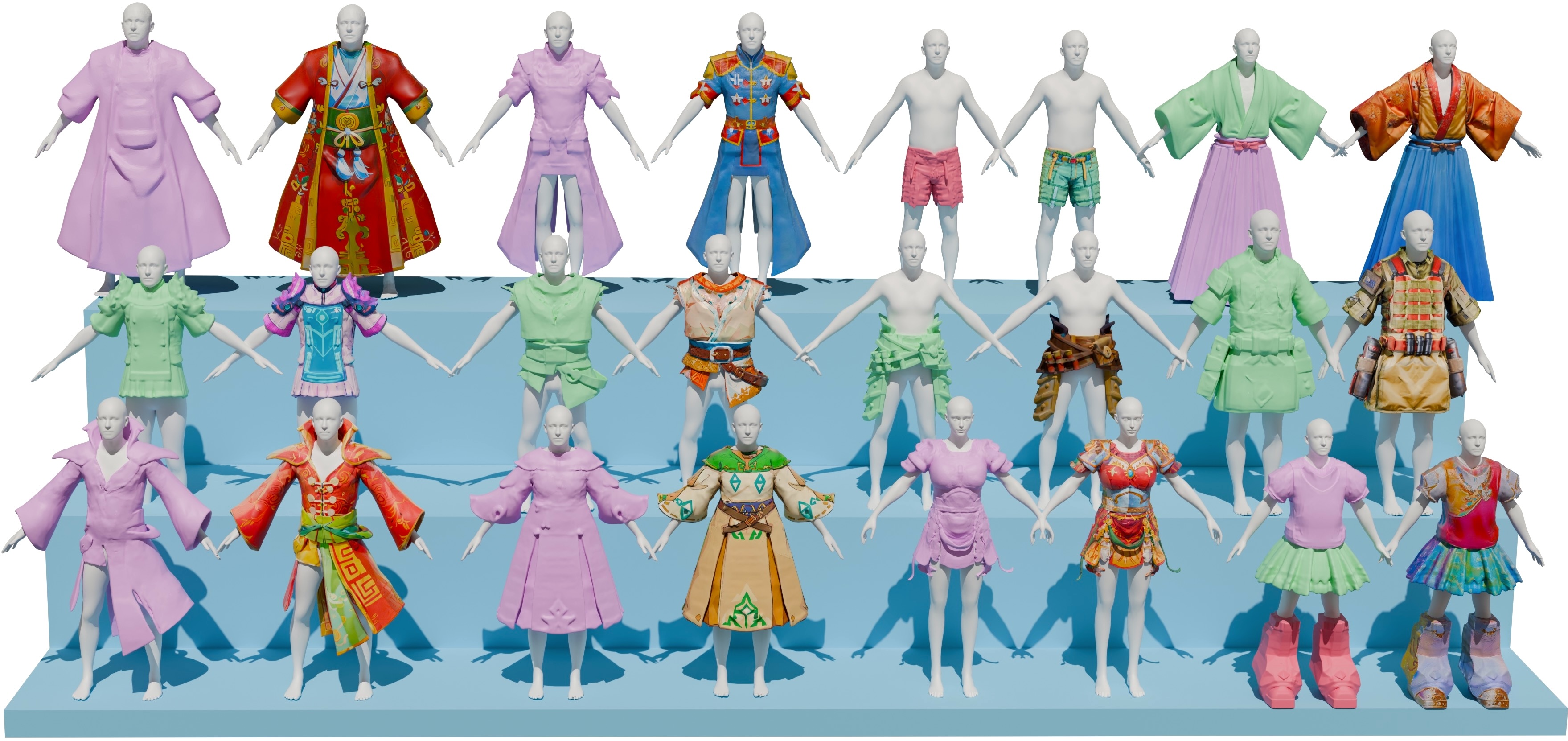}
  \caption{BAG generates a variety of body-aligned 3D shapes, ranging from individual wearable assets to combinations of multiple components.}
  \label{fig:teaser}
\end{teaserfigure}

\author{Zhongjin Luo}
\affiliation{%
 \institution{The Chinese University of Hong Kong, Shenzhen}
 \country{China}
 }
\authornote{The contribution is made during an internship at Tencent XR Vision Labs}

\author{Yang Li}
\affiliation{%
 \institution{Tencent XR Vision Labs}
 \country{China}}
\authornote{ Project lead } 

\author{Mingrui Zhang}
\affiliation{%
 \institution{Tencent XR Vision Labs}
 \country{China}
 }

\author{Senbo Wang}
 
\affiliation{%
\institution{Tencent XR Vision Labs}
\country{China}}

\author{Han Yan}
 
\affiliation{%
\institution{Shanghai Jiao Tong University}
\country{China}}
\authornotemark[1]

\author{Xibin Song}
 
\affiliation{%
\institution{Tencent XR Vision Labs}
\country{China}} 

\author{Taizhang Shang}
 
\affiliation{%
\institution{Tencent XR Vision Labs}
\country{China}}

\author{Wei Mao}
 
\affiliation{%
\institution{Tencent XR Vision Labs}
\country{China}}

\author{Hongdong Li}
 
\affiliation{%
\institution{Australian National University}
\country{Australia}}

\author{Xiaoguang Han}
 
\affiliation{%
\institution{The Chinese University of Hong Kong, Shenzhen}
\country{China}}
\authornote{Corresponding author}

\author{Pan Ji}
 
\affiliation{%
\institution{Tencent XR Vision Labs}
\country{China}}
\authornotemark[3]

\begin{abstract}
While recent advancements have shown remarkable progress in general 3D shape generation models, 
the challenge of leveraging these approaches to automatically generate wearable 3D assets remains unexplored.
To this end, we present BAG, a \textbf{B}ody-aligned \textbf{A}sset \textbf{G}eneration method to output 3D wearable asset that can be automatically dressed on given 3D human bodies.
This is achived by controlling the 3D generation process using human body shape and pose information.
Specifically, we first build a general single-image to consistent multiview image diffusion model, 
and train it on the large Objaverse dataset to achieve diversity and generalizability.
Then we train a Controlnet to guide the multiview generator to produce body-aligned multiview images.
The control signal utilizes the multiview 2D projections of the target human body, where pixel values represent the XYZ coordinates of the body surface in a canonical space.
The body-conditioned multiview diffusion generates body-aligned multiview images, which are then fed into a native 3D diffusion model to produce the 3D shape of the asset.
Finally, by recovering the similarity transformation using multiview silhouette supervision and addressing asset-body penetration with physics simulators, the 3D asset can be accurately fitted onto the target human body.
Experimental results demonstrate significant advantages over existing methods in terms of image prompt-following capability, shape diversity, and shape quality. Our project page is available at \href{https://bag-3d.github.io/}{\textcolor{blue}{https://bag-3d.github.io/}}.
\end{abstract}


\maketitle

\section{Introduction}
\label{sec:intro}

3D wearable assets, such as garments, shoes, and headwear, are vital components in the creation of digital 3D avatars. Generating a large number of high-quality 3D wearable assets is therefore crucial for many practical applications, including video games, filmmaking, and augmented and virtual reality (AR/VR).

Existing wearable asset generation is dominated by sewing pattern modeling, as seen in works like DressCode~\cite{he2024dresscode}, or PCA-based approaches, such as BCNet~\cite{jiang2020bcnet}.
While these methods offer production-ready quality shapes, they suffer from low diversity, poor prompt-following capabilities, and may fail to generate very complex geometry.

On the other hand, powered by advancements in transformers~\cite{vaswani2017attention_is_all_need} and diffusion models~\cite{ho2020denoising}, there has been substantial progress in the field of general 3D shape generation.
New methods are continuously being proposed, consistently pushing the limitations of geometric quality. 
Notable recent works include DreamFusion~\cite{poole2022dreamfusion}, LRM~\cite{hong2023lrm}, CLAY~\cite{zhang2024clay}, Tripo~\footnote{https://www.tripo3d.ai/\label{fn:tripo}} and Trellis~\cite{xiang2024trellis}, among others.
These methods excel at following input images or text prompts and can generate highly complex shapes.
By combining additional remeshing tools like~\cite{hao2024meshtron} for post-processing, 3D asset generation that meets industrial standards is becoming possible.

This brings us to the question: \textit{Can we generate body-aligned wearable 3D assets using the most advanced 3D generative models available today?}
A straightforward approach is to first generate the 3D asset and then put it onto a 3D human body model.
However, this usually requires manual intervention, such as dragging and morphing in a 3D graphics user interface, which is both time-consuming and costly,
moreover, the generated 3D assets are not guaranteed to match the shape of the target human body, which could lead to poor dressing results.

To automate the process of wearing generated 3D assets onto human models, 
Garment3DGen~\cite{garment3dgen} deforms predefined garment templates to match the generated 3D asset.
It then wraps the target human model to wear the generated asset according to the deformation of the garment template. 
This strategy is limited by the number of available template shapes. 
Additionally, the deformation usually cause severe surface stretching when the shape of the garment template significantly differs from the generated asset.

In this paper, we present an method to directly generate body-aligned 3D assets, 
such that the assets can be automatically fitted to a given human body shape without human intervention.
This is achived by controlling the 3D generation process using human body shape and pose information.
Specifically, similar to Zero123+~\cite{shi2023zero123plus}, we first build a general single-image to consistent multiview image diffusion model, 
and train it on the large Objaverse~\cite{deitke2023objaverse} dataset to achieve diversity and generalizability.
Then we train a Controlnet~\cite{zhang2023controlnet} to guide the multiview diffusion to generate body-aligned multiview images.
The control signal comprises multiview 2D projection of the target human body, with pixel values indicating the XYZ coordinates of the body surface in a canonical space.
Our body-conditioned multiview generator produce body-aligned multiview images, which are then fed into
a native 3D diffusion model to produce the 3D shape of the asset.
Finally, by optimizing a similarity transformation using multiview silhouette supervision and addressing asset-body penetration with physics simulators, the 3D asset can be accurately fitted onto the target human body.

This paper focuses on developing a method for guiding advanced 3D generative models to directly generate body-aligned 3D assets.
Our approach demonstrates significant advantages over existing methods in terms of prompt-following capacity, shape diversity, and shape quality.

\begin{figure*}[htbp]
  \centering
  \includegraphics[width=1\linewidth]{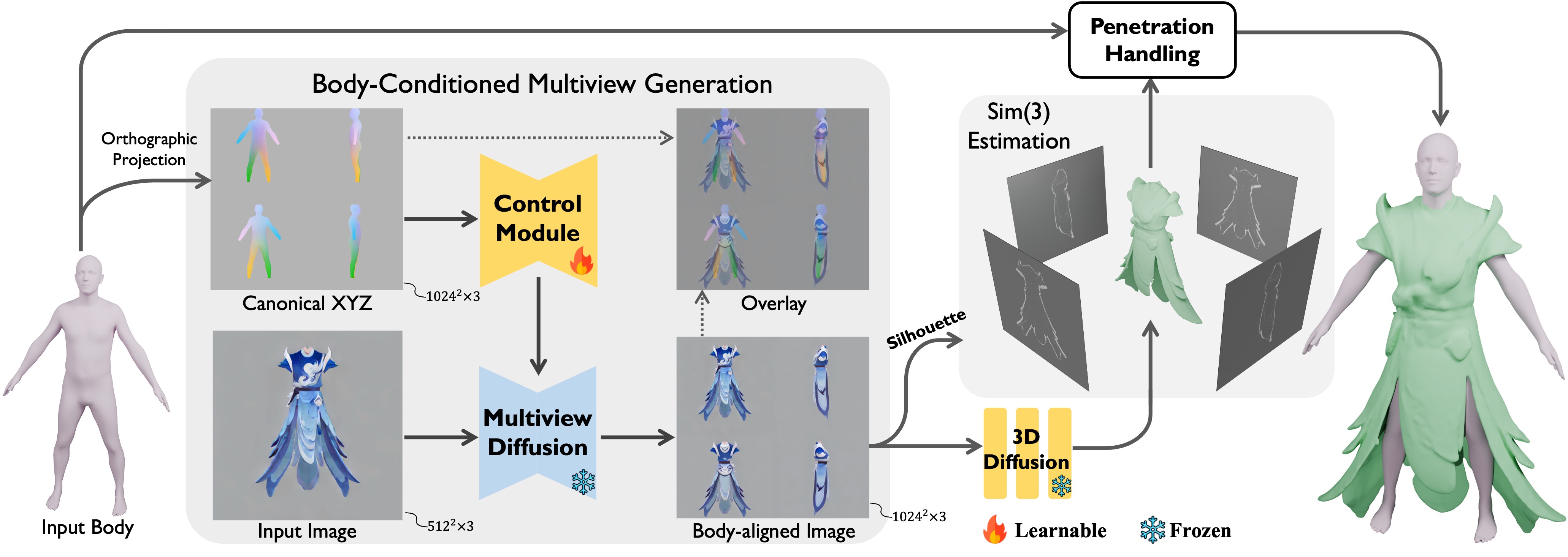}
  \caption{\textbf{Method Pipeline.}  
Given an input image and a target body, we employ body-conditioned image generation to produce body-aligned consistent four-view orthographic images (see Sec.~\ref{sec:body_0123}). 
The four-view images are then fed into a native 3D diffusion model to obtain the asset shape.
The similarity transformation (Sim3) of the generated asset is estimated through silhouette-based projection optimization (see Sec.~\ref{sec:sim3_estimation}). 
Finally, after solving the body-asset penetration, the Sim3-transformed asset is fitted onto the human body (see Sec.~\ref{sec:penetration_solve}). 
The means for obtaining the input body and image pair are detailed in Sec.~\ref{sec:input_aquisition}.
  }
  \label{fig:pipeline}
\end{figure*}

\section{Related Work}\label{sec:related}

This section surveys generative models for 3D shapes, garments, and 3D avatars. For clarity, in this paper, we regard single-view 3D reconstruction as an image-conditioned generation problem.
 
\paragraph{\textbf{General 3D Shape Generation}}~\label{Sec:native_3D_generation}
Early methods primarily leveraged Generative Adversarial Networks (GANs)~\cite{goodfellow2020GAN} to model 3D distributions, as demonstrated in 3DGAN~\cite{wu2016learning}, EG3D~\cite{EG3D} and Get3D~\cite{gao2022get3d}. However, these methods faced challenges in scaling to more diverse scenarios.
DreamFusion~\cite{poole2022dreamfusion} and its follow-up works~\cite{lin2023magic3d, chen2023fantasia3d} obtain 3D shapes by leveraging 2D generative models. However, these methods are time-consuming and suffer from the Juanus-face problem.
LRM~\cite{hong2023lrm} utilizes large transformer models to reconstruct the triplane representation of 3D shapes.
Recent native 3D generative models employ diffusion models~\cite{ho2020denoising} to generate 3D content using various representations, such as point clouds~\cite{nichol2022point-e, zeng2022lion}, meshes~\cite{Liu2023MeshDiffusion}, voxel grids~\cite{lasdiffusion, ren2024xcube, xiang2024trellis}, octrees~\cite{xiong2024octfusion}, triplanes~\cite{rodin, nfd, SSDNerf, blockfusion}, irregular primitives~\cite{chen2024primx}, vecsets~\cite{zhang20233dshape2vecset}, and network weights~\cite{erkocc2023hyperdiffusion}.
To produce shapes with artist-like topology, PolyGen~\cite{nash2020polygen} and its follow-up works~\cite{siddiqui2023meshgpt,tang2024edgerunner, weng2024bpt, hao2024meshtron} adopted autoregressive models for mesh generation.
Shape generation can follow image prompts by incorporating visual features into the generation process with transformers. One can also convert a single image to consistent multi-view images as conditioning signals using models such as Zero123+~\cite{shi2023zero123plus}. Similar techniques have been adopted in Wonder3D~\cite{long2023wonder3d}, CLAY~\cite{zhang2024clay}, Tripo~\footref{fn:tripo} and Trellis~\cite{xiang2024trellis}.
In this paper, we introduce body-conditioned multi-view image generation and leverage advanced native 3D generative models to produce 3D wearable shapes that can be automatically draped on 3D human bodies.

\paragraph{\textbf{Garment Generation}} 
Sewing patterns are a widely adopted geometric representation for generating garments, as done in Sewformer~\cite{liu2023sewformer} and DressCode~\cite{he2024dresscode}.
It is usually follow-uped with a draping process that attach cloth on human body. This can be done via physcis-based approach as done in~\cite{Baraff1998LargeSI} or data-driven approaches like DrapeNet~\cite{de2023drapenet}.
While these approaches generate production-ready mesh quality, creating sewing patterns that precisely follow image prompts remains challenging.
Some other works model garment geometry as functions from human SMPL~\cite{loper2015smpl} body pose and shape parameters,  this can be seen in CAPE~\cite{CAPE:CVPR:20}, SMPLicit~\cite{corona2021smplicit} and NeuralTailor~\cite{NeuralTailor2022}.
Among them, CAPE~\cite{CAPE:CVPR:20} regress cloth gemetry as vertex offsets from body surface mesh, SMPLicit~\cite{corona2021smplicit} generate clothes as implicit model conditioned with the SMPL human body parameters.
Garments can also be represented using vertex-based PCA models, as demonstrated in MultiGarmentNet~\cite{Multi-garment-net}, Cloth3D\cite{bertiche2020cloth3d}, BCNet~\cite{jiang2020bcnet}, and GarverseLOD~\cite{luo2024garverselod}, however, these methods are limited by the number of base garment templates.
Garment3DGen~\cite{garment3dgen} leverages Wonder3D to generate a reference garment shape and deforms a selected garment template to fit the generated shape, however, this often causes surface stretches when the reference shape significantly differs from the template shapes.
The aforementioned methods usually suffer from poor prompt-following ability, lack of diversity in generated shapes, and failure to generate complex geometric structures. Recent native 3D generative models surveyed in Sec.~\ref{Sec:native_3D_generation} offer high diversity and good prompt-following capacity. In this paper, we leverage native 3D generative models to produce 3D wearable shapes that can be automatically draped on 3D human bodies.

\paragraph{\textbf{3D Avatar Generation}}
Early works on 3D avatar generation~\cite{bergman2022gnarf,chen2022gdna,noguchi2022unsupervised} could generate diverse human geometries or textured avatars, but the generation process could not be controlled by text descriptions.
Text-conditioned avatar generation is made possible with the help of large language models like CLIP~\cite{clip} and text-to-image diffusion models~\cite{stable_diffusion}.
AvatarCLIP~\cite{hong2022avatarclip} leverage CLIP space loss to  supervise neural human generation, including 3D geometry, texture and animation.
Several works extend the Score Distillation Sampling (SDS) method~\cite{poole2022dreamfusion} to generate avatars with the guidence of human SMPL~\cite{loper2015smpl} model, notable works includes Dreamavatar~\cite{cao2024dreamavatar}, TADA~\cite{liao2024tada}, and HumanNorm~\cite{huang2023humannorm}. 
Image-conditioned avatar generation is closely related to human reconstruction, as seen in  PIFu~\cite{saito2019pifu}, PAMIR~\cite{zheng2021pamir}, and ICON~\cite{xiu2022icon}. 
These methods usually involve single-view human body shape and pose estimation, followed by holistic shape reconstruction.
CharacterGen~\cite{peng2024charactergen} controls avatar generation using human body poses.
The aforementioned approaches generate avatars as a single entangled shape, whereas downstream applications require disentangled shapes, such as separate hair, clothing, and bodies.
Reconstructing the human body and clothes as separate geometries has long been studied in the field of computer vision. Notable works include ClothCap~\cite{Pons-Moll:Siggraph2017ClothCap} and Delta~\cite{Feng2023DELTA}.
Layered avatar generation can be achieved through separate SDS optimization processes for the human body, clothing, and hair, respectively. This technique is demonstrated in TECA~\cite{zhang2023teca} and SOSMPL~\cite{wang2023sosmpl}.
Frankenstein~\cite{frankenstein} generates multiple parts of an avatar from a single triplane reverse diffusion process. This is achieved by compressing the layered avatar shape into a multi-head triplane representation.
STDGen~\cite{he2024stdgen} adopts a similar idea with the LRM architecture.

\section{Method}
\label{sec:method}

The overview of the method is shown in Fig.~\ref{fig:pipeline}.

\subsection{Body Conditioned Multi-view Image Generation}~\label{sec:body_0123}

This section outlines the method for constructing a body-aligned multi-view consistency image generation module. 
This is achived by controlling the 3D generation process using human body shape and pose information.

\medskip
\noindent
\textbf{Consistent Multiview Image Generation.} 
First, we build a general single-image to consistent multiview image diffusion model similar to Zero123++~\cite{shi2023zero123plus}.
The multi-view images are represented by tiling four images in a $2\times2$ layout to form a single frame.
The four training views are rendered using orthographic cameras, maintaining an absolute elevation angle of \(0^\circ\) and four fixed azimuth angles: \{ \(0^\circ\), \(90^\circ\), \(180^\circ\), \(270^\circ\) \}. Therefore, the first view corresponds to the input front view, while the remaining views represent the relative left, back, and right perspectives of the object.
Similar to Zero123++, we reuse the weights of Stable Diffusion 2.0~\cite{stable_diffusion} and finetune the weigths using rendered images of over 600k objects filtered from the original Objaverse~\cite{deitke2023objaverse} dataset.

\begin{figure}[htbp]
  \centering
  \includegraphics[width=1.08\linewidth]{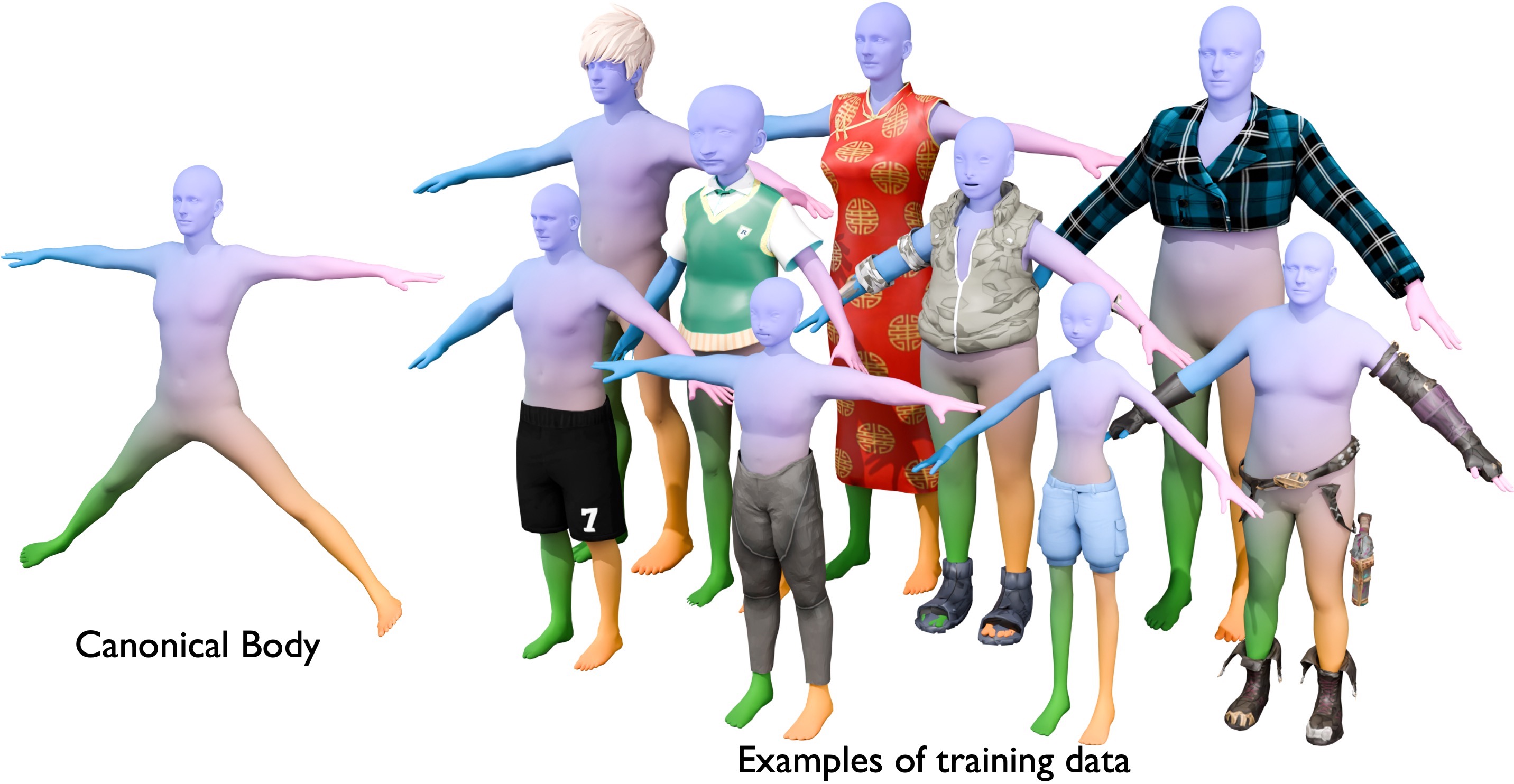}
  \caption{Canonical body space (left), and examples of body-aligned 3D asset dataset (right). The color on the body surface is obtained by scaling the canonical XYZ values to the range of [0-255].}
  \label{fig:data_example}
\end{figure}

\medskip
\noindent
\textbf{Body-Aligned Multiview-Generation.} 
Now that we have a general multiview generator, we can control it to generate body-aligned multiview images.
First, we constructed a dataset of body-aligned 3D assets, including hair, clothing, trousers, outfits, and shoes, etc. 
Examples of these assets are illustrated in Fig.~\ref{fig:data_example}.
Each 3D asset is associated with a human SMPLX body~\cite{SMPL-X:2019}. 
The SMPLX shape parameters are randomly sampled from the SMPLX Betas space. 
The pose parameters are randomly interpolated between a T-pose and an A-pose.
We retarget the assets to the randomly sampled SMPLX bodies, ensuring that the assets align well with the bodies in 3D space.
The retargeting technique is  based on ClothCap~\cite{Pons-Moll:Siggraph2017ClothCap}.
As depicted in Fig.~\ref{fig:pipeline}, we employ a consistent orthographic camera to project the SMPLX mesh onto tiled four-view images, which serve as body conditioning signals. 
Each pixel value in these body images corresponds to the XYZ coordinates in a fixed canonical body shape, as shown in Fig.~\ref{fig:data_example}. 
The coordinate values are normalized to the range [0, 1].
Since the 3D assets are body-aligned, the overlay of multiview projections of the body and the assets also aligns.
In total, we obtained 70485 asset-body pairs, and we split them to train/test set with the ratio of 9/1.
We use the dataset to control the Multiview-Generation  
We train a ControlNet~\cite{zhang2023controlnet} to condition on the XYZ coordinate maps, enabling the generation of body-aligned multiview images.
The body-aligned multiview images are then input into a native 3D diffusion model to generate the asset shape.

\subsection{Similarity Transformation Estimation}~\label{sec:sim3_estimation}

Given body-aligned multiview images, we can leverage multiview-conditioned native 3D diffusion models for asset shape generation, e.g., CLAY~\cite{zhang2024clay} and Tripo~\footref{fn:tripo}.
This paper primarily focuses on \textit{body-aligned} geometry generation, the textures of the assets are generated using off-the-shelf texturing tool Meshy~\footnote{https://www.meshy.ai/}.
Due to the non-deterministic nature of the diffusion model, the shape generator cannot guarantee that the output shape will align with the input multi-view images.
This misalignment can manifest as differences in scale, translation, and rotation.

To address this issue, we optimize the similarity transformation 
$ \scalemath{0.75}{
\begin{pmatrix}
s\mathbf{R} & \mathbf{t} \\
0 & 1 
\end{pmatrix}
} \in \Sim(3)
$ with multiview image supervision, where $s\in \mathbb{R}^+$ is the scaling factor, $\mathbf{R}\in \SO(3)$ denote rotation, and $\mathbf{t}\in\mathbb{R}^3$ is translation. 
We parameterize rotation with a 3-dimensional axis-angle vector $\boldsymbol{\omega}\in\mathbb{R}^3$.
We use the exponential map 
$\exp : \mathfrak{so}(3)\rightarrow \SO(3), \quad \widehat{\hspace{0pt}\boldsymbol{\omega}\hspace{0pt}} \mapsto \mathrm{e}^{\widehat{\hspace{0pt}\boldsymbol{\omega}\hspace{0pt}}} = \mathbf{R}$ 
to convert from axis-angle to matrix rotation form, 
where the~$\, \widehat{\cdot}$-operator creates a $3 \times 3$ skew-symmetric matrix from a 3-dimensional vector. 
Given an generated asset mesh with vertex set $\mathbf{V}_{\text{raw}}=\{\mathbf{x}_i\in\mathbb{R}^3|_{i=1..,n}\}$,  the transformed vertex set $\mathbf{V}_{\text{transformed}}=\{\mathbf{y}_i\in\mathbb{R}^3|_{i=1..,n}\}$ are obtained via,
\begin{equation}
  \mathbf{y}_i = s\mathrm{e}^{\widehat{\hspace{0pt}\boldsymbol{\omega}\hspace{0pt}}} \mathbf{x}_i
 + \mathbf{t}
\end{equation}
We employ silhouette loss as supervision, which is the pixel-wise mean absolute difference between the rendered multiview silhouettes $\mathbf{S}_{\text{rendered}}$ and the input multiview silhouettes $\mathbf{S}_{\text{input}}$. 
The loss reads
\begin{equation}
\mathcal{L}_{\text{silhouette}} = \frac{1}{N} \sum_{i=1}^N \left| \mathbf{S}_{\text{rendered}, i} - \mathbf{S}_{\text{input}, i} \right|.
\end{equation}
Where $N$ is the total number of pixels.
The input silhouettes are obtained by running SAM~\cite{kirillov2023SAM}. 
The silhouettes of the mesh are rendered using the Phong shader implemented in PyTorch3D.
Optimization is performed using the Adam optimizer.

\subsection{Handling Asset-Body Penetration}~\label{sec:penetration_solve}

The Sim(3)-transformed asset can roughly align with the body, but small penetrations between the asset and body still exist. To address this issue, we further resolve such penetrations using a physics simulation-based solver. 

\begin{figure}[htbp]
  \centering
  \hspace*{-0.5cm}
  \includegraphics[width=1.1\linewidth]{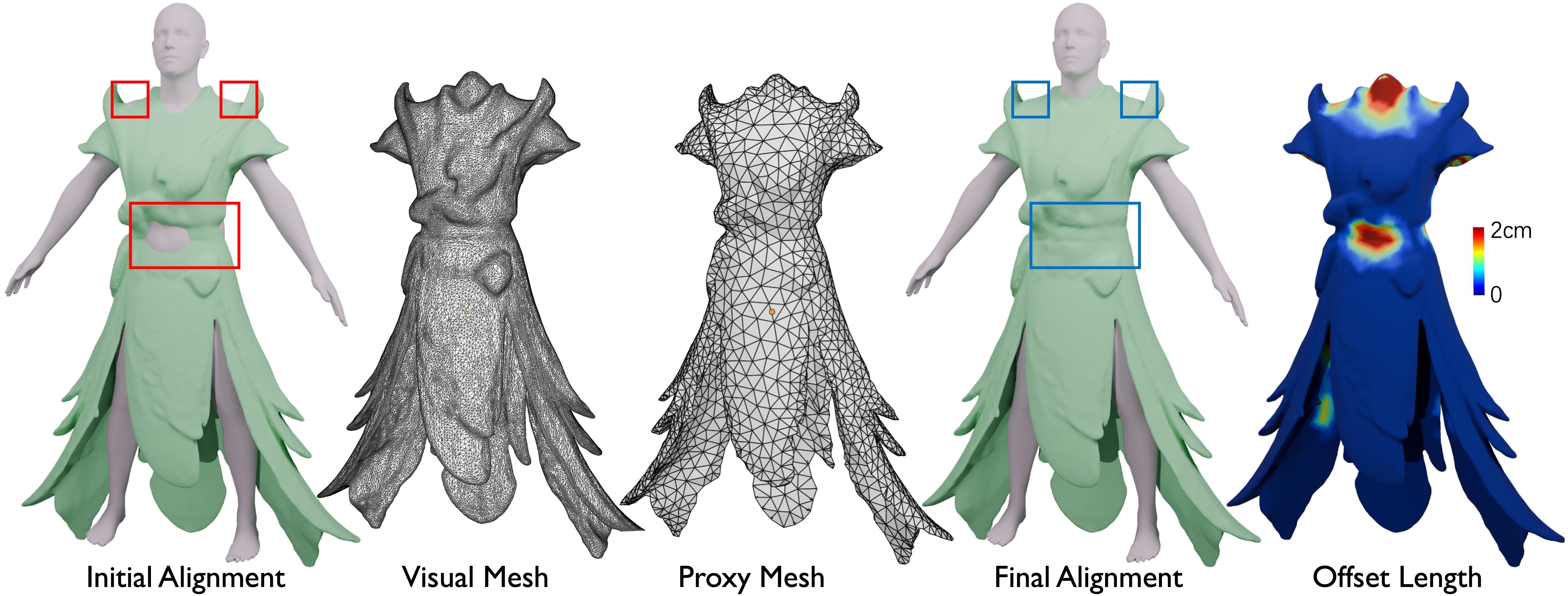}
  
  \caption{\textbf{Penetration Handling.} Despite the application of the Sim(3) transformation, penetrations between the asset and the body persist, as illustrated in the Initial Alignment. To address this, a proxy mesh is employed, which retains the essential geometry of the visual mesh and serves as a representative for cloth simulation. The Final Alignment showcases the penetration-free state of the asset and body post-simulation.}
  \label{fig:pen_solve}
\end{figure}

\medskip
\noindent
\textbf{Representing Asset via Single Layer Proxy Mesh.} 
Our native 3D diffusion model represents 3D geometry as occupancy~\cite{occupancy_networks} and extracts the surface mesh using the marching cubes algorithm. As a result, the generated assets are watertight but not suitable for physics simulation. For instance, the generated garments are not single-layer meshes; they have thickness, which can cause issues when simulating body-asset collisions. 
Inspired by ProxyCloth~\cite{ProxyCloth}, we use single-layer proxy mesh to represent the asset.
Specifically, we first put cameras around the human body and remove triangles that can not be observed by the cameras.
Then we simplifiy the mesh with Voronoi diagram-based technique~\cite{valette2008Voronoi}, which clusters the vertices in into a Centroidal Voronoi Diagram and then extract the proxy from the centroids of neighboring clusters.
After removing non-manifold faces and non-manifold vertices, we obtain a uniformly meshed surface.
An example of comparison bewteen original visual mesh and proxy mesh is shown in Fig.~\ref{fig:pen_solve}.
Then we compute the Linear Blend Skinning (LBS) weights from the asset shape to asset proxy, which can propagate the potential deformation back to the visual mesh. 

\medskip
\noindent
\textbf{Penetration Handling.} To physically resolve penetration between the asset and the body, we developed a position-based simulation of compliant constrained dynamics (XPBD)~\cite{Macklin2016XPBD} based cloth simulator. The simulator can hanlde the penetration through collision response between the asset and the body.
We compute a Signed Distance Field (SDF) for the body to facilitate collision detection. During each simulation step, vertices on the asset with negative SDF values are identified as penetrated into the body. These vertices are projected to a penetration-free state based on their SDF normals and values using a distance-based constraint. Once penetration is resolved, a frictional position constraint is applied to account for surface interaction.
Additionally, a vertex-to-vertex distance-based collision constraint is employed to manage asset self-collisions. To enhance performance, we use spatial hashing~\cite{teschner2003optimized} to accelerate the neighbor search process for vertices.
The deformed, penetration-resolved asset is treated as the updated rest state. To minimize deformation and preserve the asset’s shape, gravity forces are omitted during penetration handling.
The simulation is performed on the single-layer proxy mesh previously mentioned, and the resulting deformations are propagated back to the original mesh using precomputed LBS weights.

\subsection{ Input Body And Image Pair Acquisition}~\label{sec:input_aquisition}
\begin{figure}[t]
  \centering
  \includegraphics[width=0.92\linewidth]{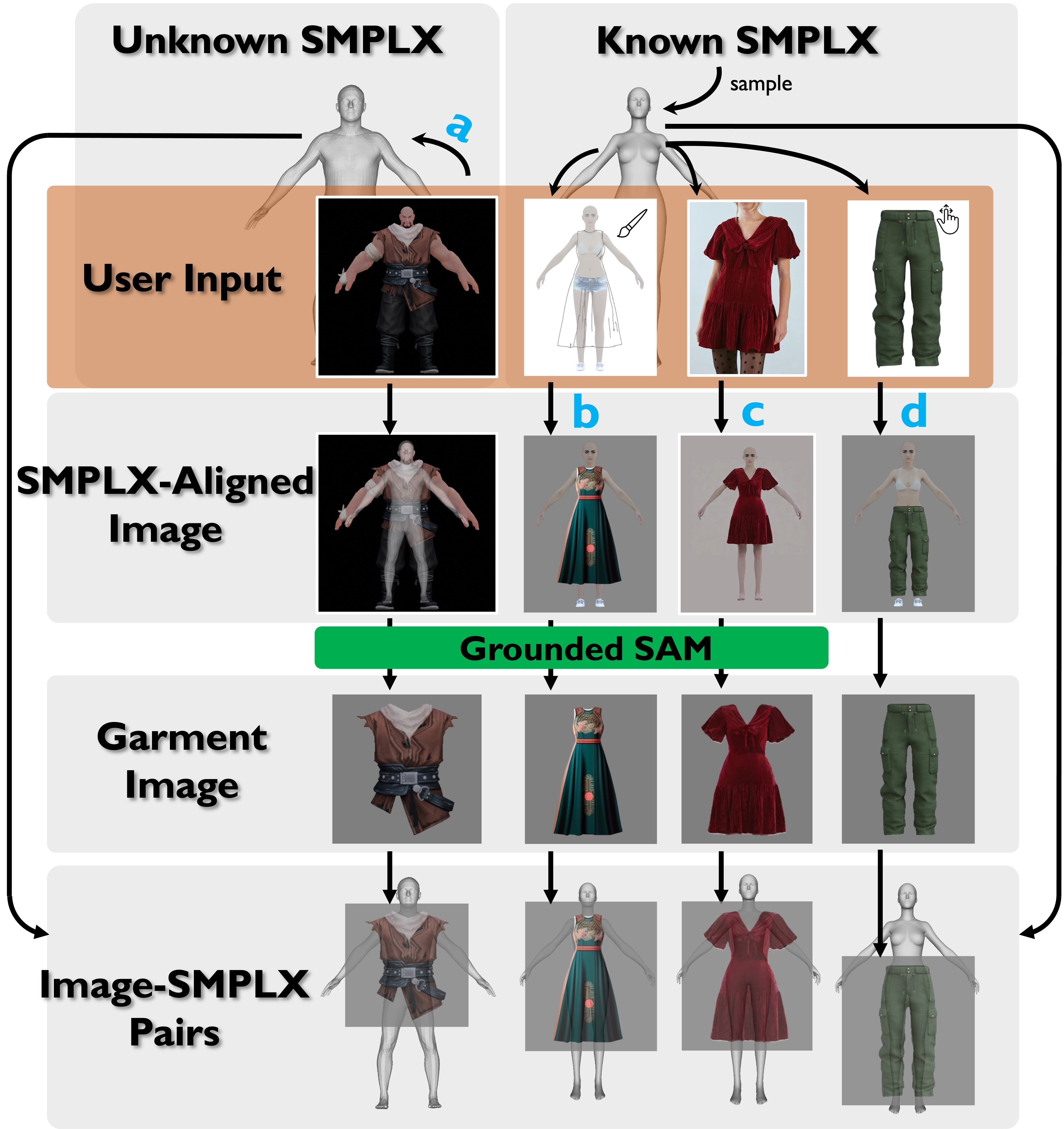}
  \caption{
  Four Methods to acqure input body and image pairs. \textbf{a}) SMPLX Fitting. \textbf{b})Sketch-Based Modeling. \textbf{c}) Virtual Try-on. \textbf{d}) Manual Images Assembly.  
  }
  \label{fig:input_acquisition}
\end{figure}

Our method requires that the input image of the asset is aligned with the front view of the body.
As shown in Fig.~\ref{fig:input_acquisition}, we identified four use cases to obtain these input pairs. The following are the details:

\smallskip
\noindent
\textbf{a) Image-based SMPLX Fitting.}
Given an image of a dressed human, we estimate the SMPLX parameters using the techniques introduced in PyMAF~\cite{pymaf2021,pymafx2023, xiu2022icon}.

\smallskip
\noindent
\textbf{b) Sketch-Based Modeling.}  
Users can draw 2D assets on top of the rendered image of a SMPLX model. The sketch is then converted into a colored image using SDXL~\cite{stable_diffusion} and T2i-adaptor~\cite{mou2023t2i}
based sketch-to-image generation.

\smallskip
\noindent
\textbf{c) Image-Based Virtual Try-On (VTON).}  
VTON takes an image of a garment as input and generates an outfitted image of a target human wearing the garment.
We leverage Kolors~\cite{kolors}~\footnote{https://huggingface.co/spaces/Kwai-Kolors/Kolors-Virtual-Try-On} as our VTON tool. 
We use the rendered image of a textured SMPLX as the target 2D human. 

\smallskip
\noindent
\textbf{d) Manual Image Assembly.} 
Users can manually drag images of existing 2D assets onto the rendered image of a textured SMPLX model.
Note that only case \{\textbf{a}\} requires human pose estimation; in cases \{\textbf{b}, \textbf{c}, \textbf{d}\}, the body parameters are provided.
In cases \{\textbf{a}, \textbf{b}, \textbf{c}\}, the image of the asset needs to be segmented, for which we utilize Grounded SAM~\cite{ren2024grounded-sam}.
\section{Experiments}
\label{sec:experiments}

\medskip
\noindent
\textbf{Evaluation metric.}
The quality of shape generation is evaluated using the Chamfer Distance (CD), Normal Consistency (NC), and the average Point-to-Surface Euclidean distance (P2S).
The multiview image generation is evaluated with Peak Signal-to-Noise Ratio (PSNR), Structural Similarity (SSIM), and the Learned Perceptual Image Patch Similarity (LPIPS)~\cite{zhang2018LPIPS} metric.

\begin{figure*}[htbp]
  \centering
  \includegraphics[width=0.99\linewidth]{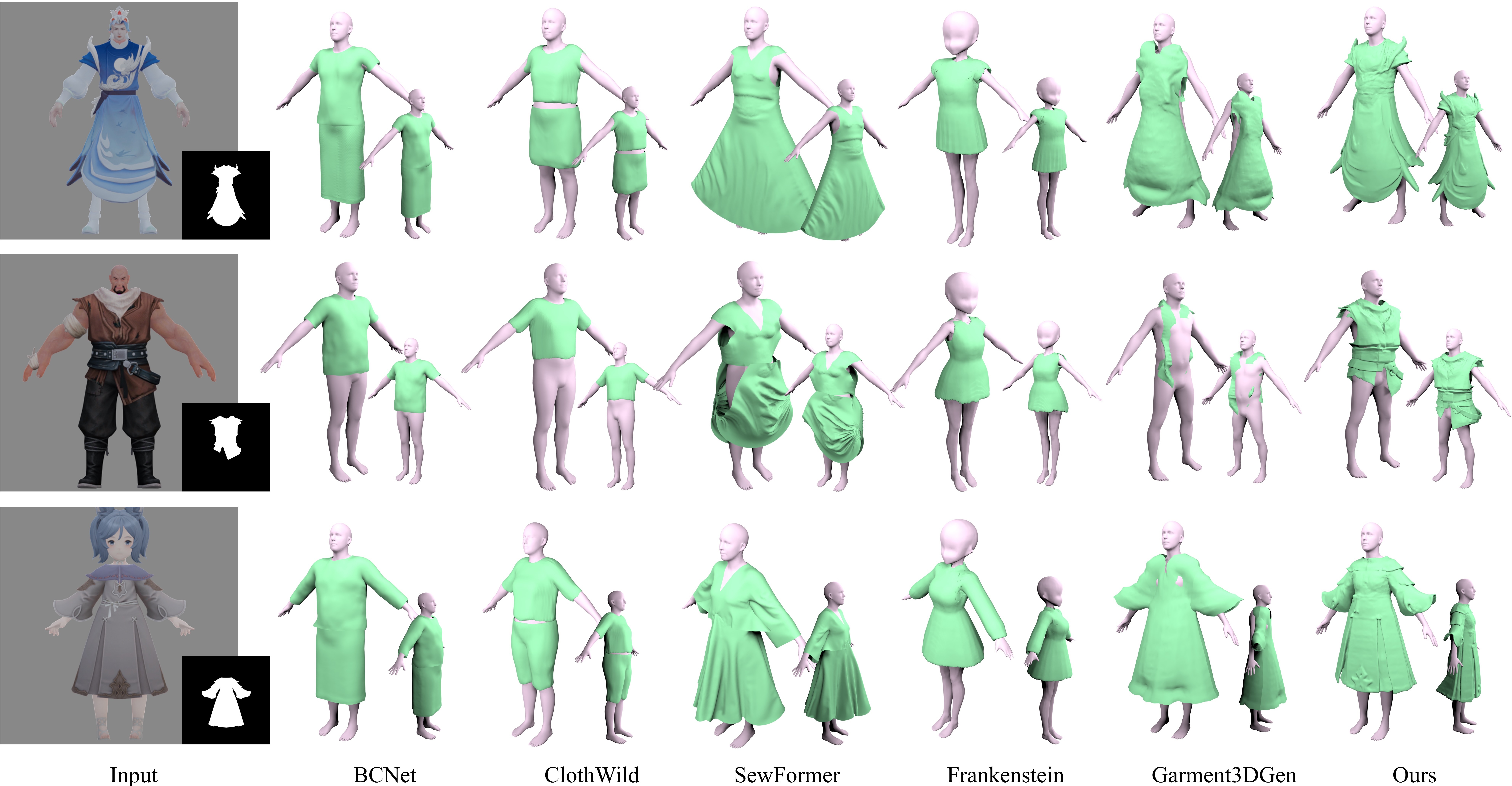}
  \caption{Qualitative asset shape generation results.
  }
  \label{fig:compare}
\end{figure*}

\begin{table}[htbp]
    \centering
    \caption{
    Quantitative Shape Generation Results.
    }
    
    \resizebox{0.7\linewidth}{!}{
        \begin{tabular}{l|cccc}
          Method & CD $\downarrow$ & NC $\uparrow$ & P2S $\downarrow$ \\
        \midrule
        BCNet       & 21.031 & 66.728 & 13.673  \\
        ClothWild   & 18.995 & 70.238 & 11.711  \\
        SewFormer     & 23.945 & 66.467 & 15.627  \\
        Frankenstein  & 20.996 & 68.975 & 12.707 \\
        Garment3DGen & 17.347 & 72.687 & 9.816  \\
        Ours         & \textbf{9.688}  & \textbf{79.361} & \textbf{6.398}  \\
        \end{tabular}
    }
    \label{tab:compare}
\end{table}

\begin{figure*}[htbp]
  \centering
  \includegraphics[width=0.89\linewidth]{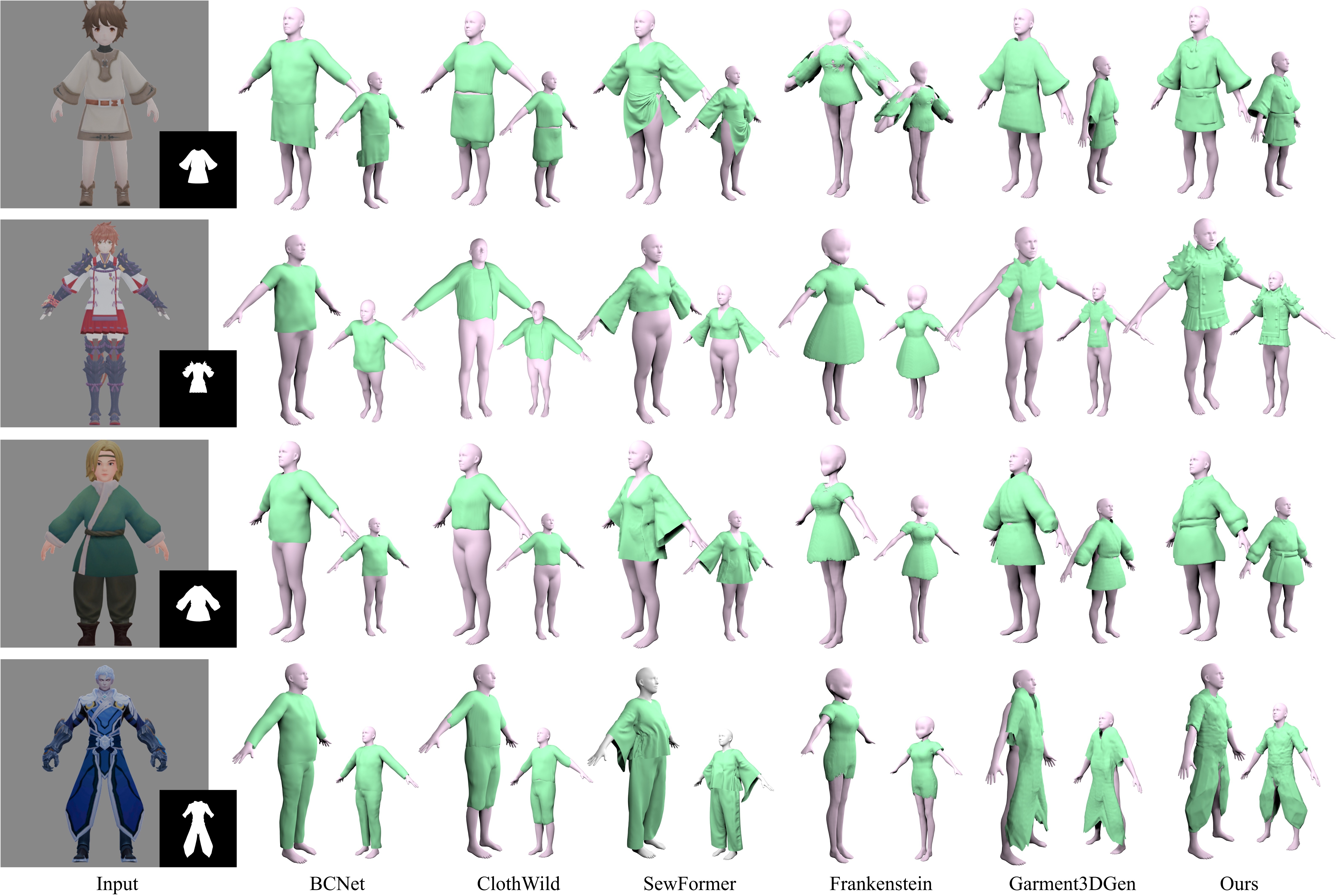}
  \caption{Qualitative comparison between ours and the state of the arts. For each row, the input image is followed by the results generated by BCNet~\cite{jiang2020bcnet}, ClothWild~\cite{Moon_2022_ECCV_ClothWild}, SewFormer~\cite{liu2023sewformer}, Frankenstein~\cite{yan2024frankenstein}, Garment3DGen~\cite{garment3dgen} and our method.}
  \label{fig:compare_more}
\end{figure*}

\medskip
\noindent
\textbf{Comparison with SOTA.} 
We compare our method with the state-of-the-art (SOTA) single-view garment reconstruction methods, including BCNet~\cite{jiang2020bcnet}, ClothWild~\cite{Moon_2022_ECCV_ClothWild},SewFormer~\cite{liu2023sewformer}, Frankenstein~\cite{yan2024frankenstein}, and Garment3DGen~\cite{garment3dgen}, both quantitatively and qualitatively. 
Tab.~\ref{tab:compare} shows the quantitative results. Our method achieves the best scores against the baselines. 
Fig.~\ref{fig:compare} and Fig.~\ref{fig:compare_more} provide qualitative results of single-view image conditioned generation.
We repose the results of all the methods to the A-pose for better visualization. 
As observed, BCNet, ClothWild and SewFormer fail to generate results that align with the input image.
This limitation stems from their sole reliance on garment parametric models (whether explicit, implicit, or based on sewing patterns).
The limited representational capacity of these parametric models hinders their ability to recover aligned garment shapes that accurately follows the input images. 
Although Frankenstein employs a more effective 3D triplane representation, its stringent requirements for training data make it difficult to scale up. 
The limited availability of data results in inadequate generalization capabilities of the trained model, hindering its ability to effectively reconstruct complex garments from images. 
By harnessing the advanced generalization capabilities of large 3D models, Garment3DGen can reconstruct relatively aligned 3D shapes from images. 
However, due to the lack of prior knowledge about body shapes, the generated 3D garments are challenging to fit onto the human body, which necessitates labor-intensive and time-consuming manual post-processing to achieve body-garment alignment. 
Furthermore, its template-fitting method restricts topological flexibility, obstructing the recovery of complex structures and fine-grained geometric details. 
Our method demonstrates proficiency in effectively producing body-aligned garment shapes and faithfully capturing high-fidelity geometric details from single-view images.

\begin{figure*}[htbp]
  \centering
  \includegraphics[width=1\linewidth]{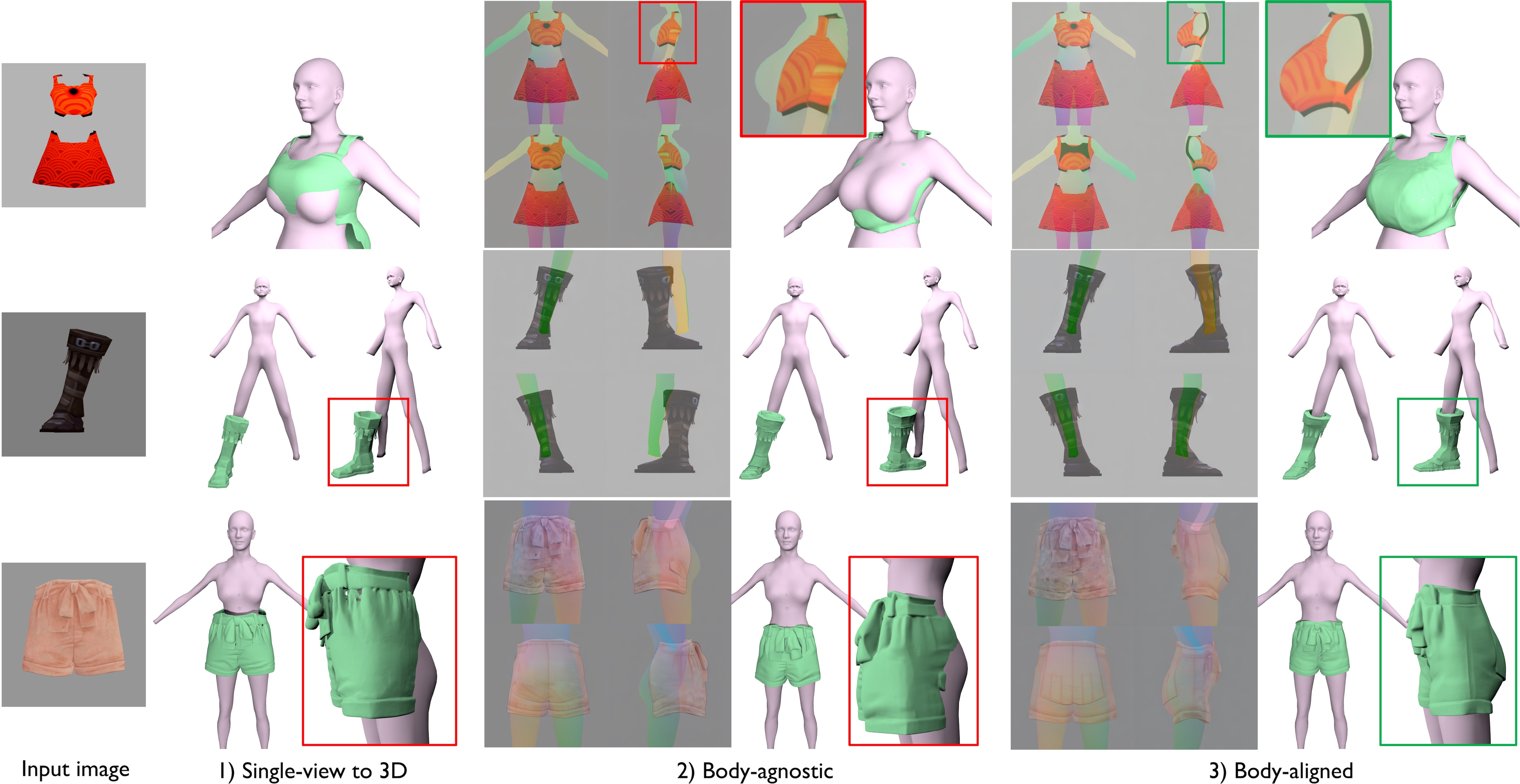}
  \caption{Ablation study of body-aligned multi-view generation.}
  \label{fig:body_condition}
\end{figure*}

\begin{table}[htbp]
    \centering
    \caption{\textbf{Quantitative comparison between our method and alternative strategies}.}
    \resizebox{0.88\linewidth}{!}{
        \begin{tabular}{c|cc|cc|cc}
        \multirow{2}{*}{Method} & \multicolumn{2}{c|}{PSNR$\uparrow$} & \multicolumn{2}{c}{SSIM$\uparrow$} & \multicolumn{2}{c}{LPIPS$\downarrow$} \\
        & Front & All & Front & All & Front & All \\
        \midrule
        w/o body & 27.168 & 17.333 & 0.733 & 0.616 & 0.0935 & 0.161  \\
        w body & \textbf{32.296} & \textbf{26.569} & \textbf{0.786} & \textbf{0.707} & \textbf{0.0893} & \textbf{0.103}  \\
        \end{tabular}
    }
    \label{tab:body_condition}
\end{table}

\begin{figure*}[htbp]
  \centering
  \includegraphics[width=0.86\linewidth]{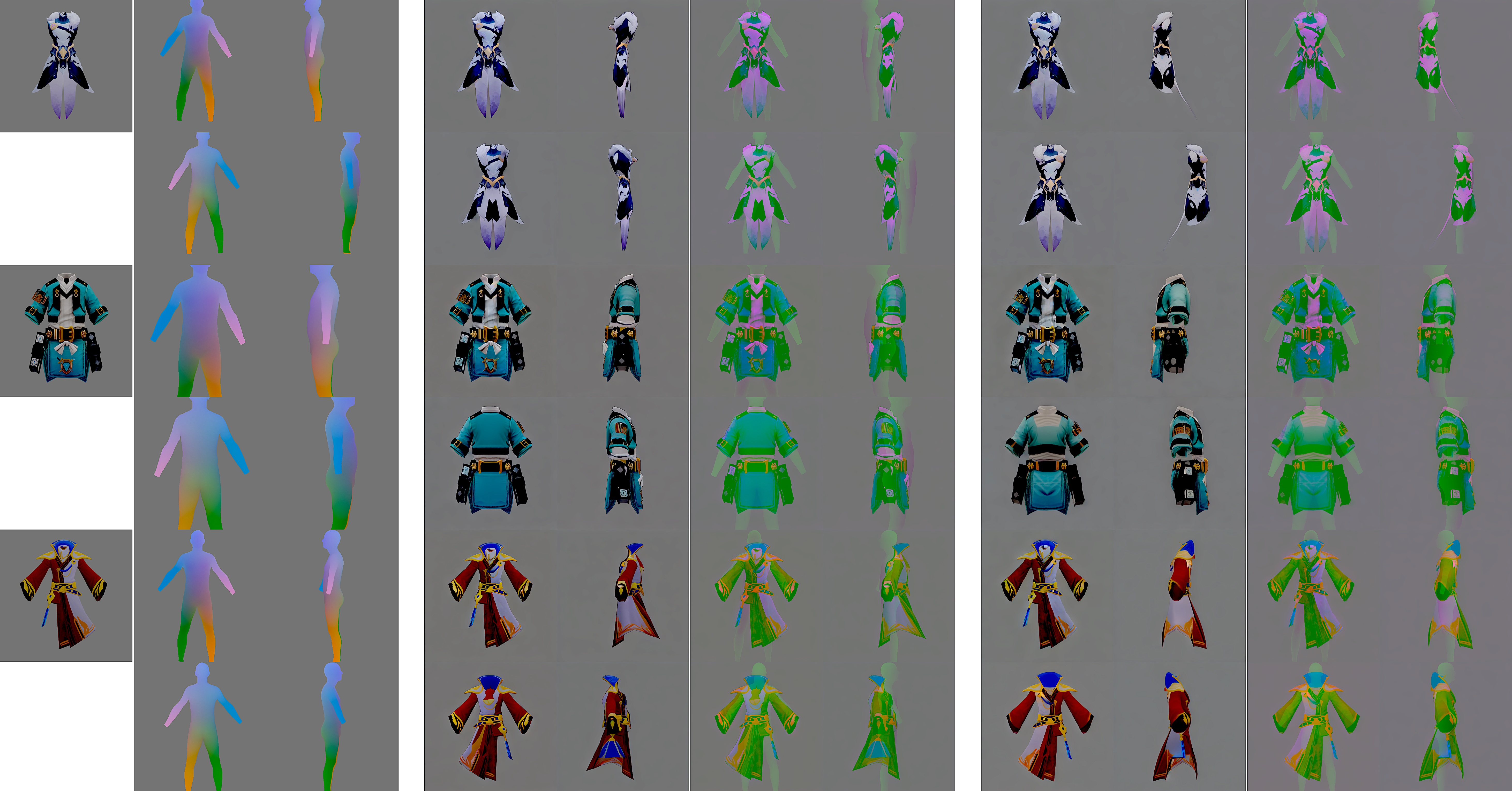}
  \caption{Qualitative comparison on multi-view generation. For each case, the input image and body condition are followed by the results generated by body-agnostic and body-aligned multi-view generation.}
  \label{fig:body_condition2}
\end{figure*}

\medskip
\noindent
\textbf{Ablation Study on body-aligned multi-view generation.}
To demonstrate the effectiveness of our body-conditioned multi-view image generation, we conduct an ablation study on various methods for obtaining 3D assets: 1) single-view to 3D asset; 2) single-view to original multi-view to 3D asset (i.e., body-agnostic); and 3) single-view to body-aligned multi-view to 3D asset. Fig.~\ref{fig:body_condition} presents comparisons of these three strategies. As observed, both the single-view and body-agnostic multi-view approaches result in body-garment misalignment. The lack of depth information in the single-view to 3D asset method often leads to discrepancies in depth between the body and the asset. The absence of body conditioning in the original multi-view generation yields multi-view images that misalign with the corresponding 2D body renderings, causing noticeable offsets between the generated multi-view images and the body renderings. In contrast, our body-aligned multi-view image generation guarantees precise alignment of assets with the human body across multiple perspectives, ensuring that the final generated 3D assets are consistently aligned with the 3D body. Tab.~\ref{tab:body_condition} presents the quantitative comparisons between body-agnostic and body-aligned multi-view image generation, which quantitatively demonstrate that our body-aligned generation performs significantly better and can produce multi-view images aligned with the human body across multiple 2D perspectives. Fig.~\ref{fig:body_condition2} presents more comparisons between body-agnostic and body-aligned multi-view generation.

\begin{figure*}[htbp]
  \centering
  \includegraphics[width=0.92\linewidth]{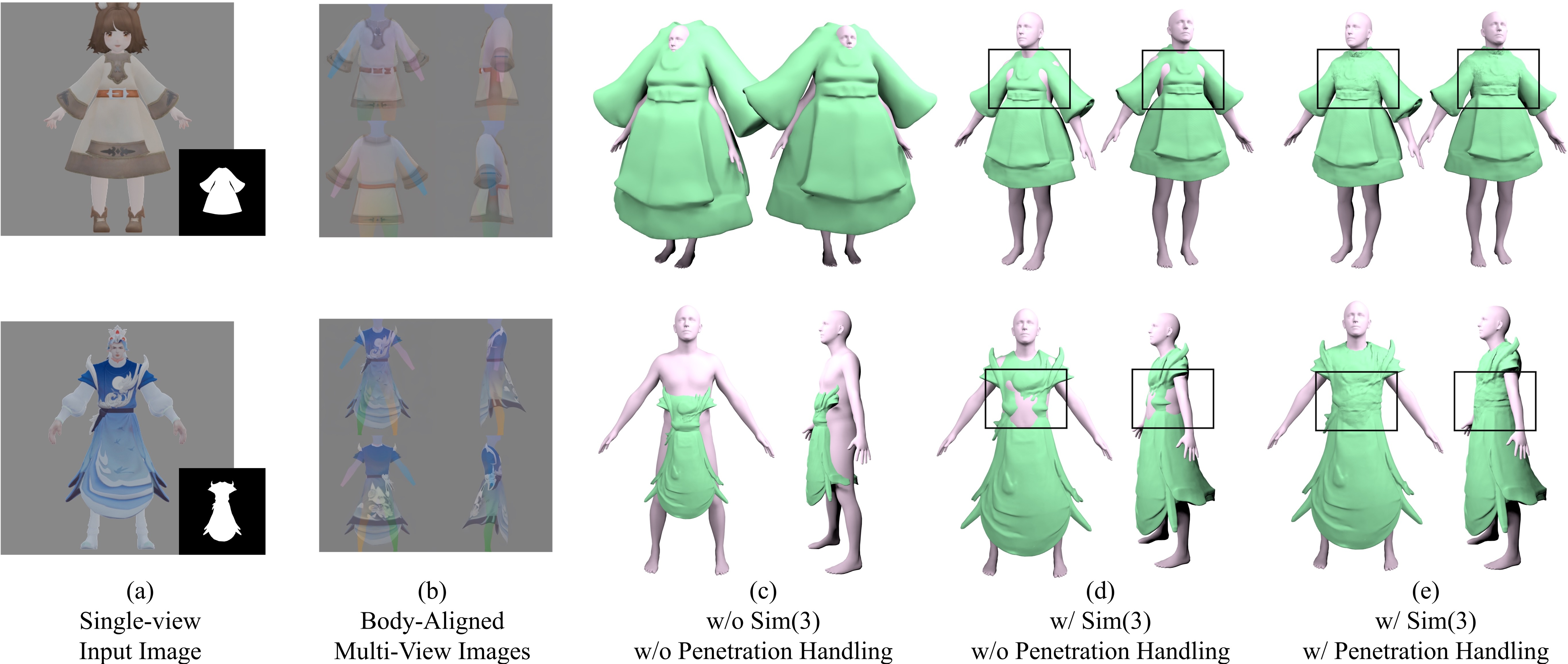}
  \caption{Qualitative comparison between ours and the ablated alignement strategies. As shown, while the generated multi-view images align with the rendered body from various 2D perspectives (b), the generated 3D assets do not effectively guarantee alignment with the input images (c). This discrepancy stems from different techniques and normalization strategies used in multiview-to-3D diffusion models. By employing $\Sim(3)$ optimization, the 3D mesh aligns with the human body (d), though some penetrations remain. Our penetration-handling approach, as illustrated in (e), achieves a penetration-free state for the asset and the body.}
  \label{fig:alignment}
\end{figure*}

\medskip
\noindent
\textbf{Ablation Study on Alignment Strategy.}
To demonstrate the significance of our asset-body alignment strategy, we conduct an ablation study on $\Sim(3)$ optimization and penetration handling. Fig.~\ref{fig:alignment} presents the comparisons between our strategy and the ablated strategies.
As demonstrated, although the generated multi-view images are aligned with the rendered body from multiple 2D perspectives (Fig.~\ref{fig:alignment}b), the 3D assets produced from these multi-view images do not effectively guarantee alignment with the input images (Fig.~\ref{fig:alignment}c). This discrepancy arises from various multiview-to-3D diffusion models employing different techniques and normalization strategies during training to accommodate their specific architectures and objectives. As a result, the generated 3D meshes are often not perfectly aligned with the input multi-view images. This misalignment manifests as differences in scale, translation, and rotation between the reconstructed 3D mesh and the input multi-view images (Fig.~\ref{fig:alignment}c). By employing $\Sim(3)$ optimization, the 3D mesh becomes geometrically consistent with the multi-view observations and consequently aligns with the human body (Fig.~\ref{fig:alignment}d); however, there are still some penetrations between the asset and the body. As seen in Fig.d, by applying our penetration handling approach, we can achieve a penetration-free state for the asset and the body (Fig.~\ref{fig:alignment}e).

\begin{figure*}[htbp]
  \centering
  \includegraphics[width=1\linewidth]{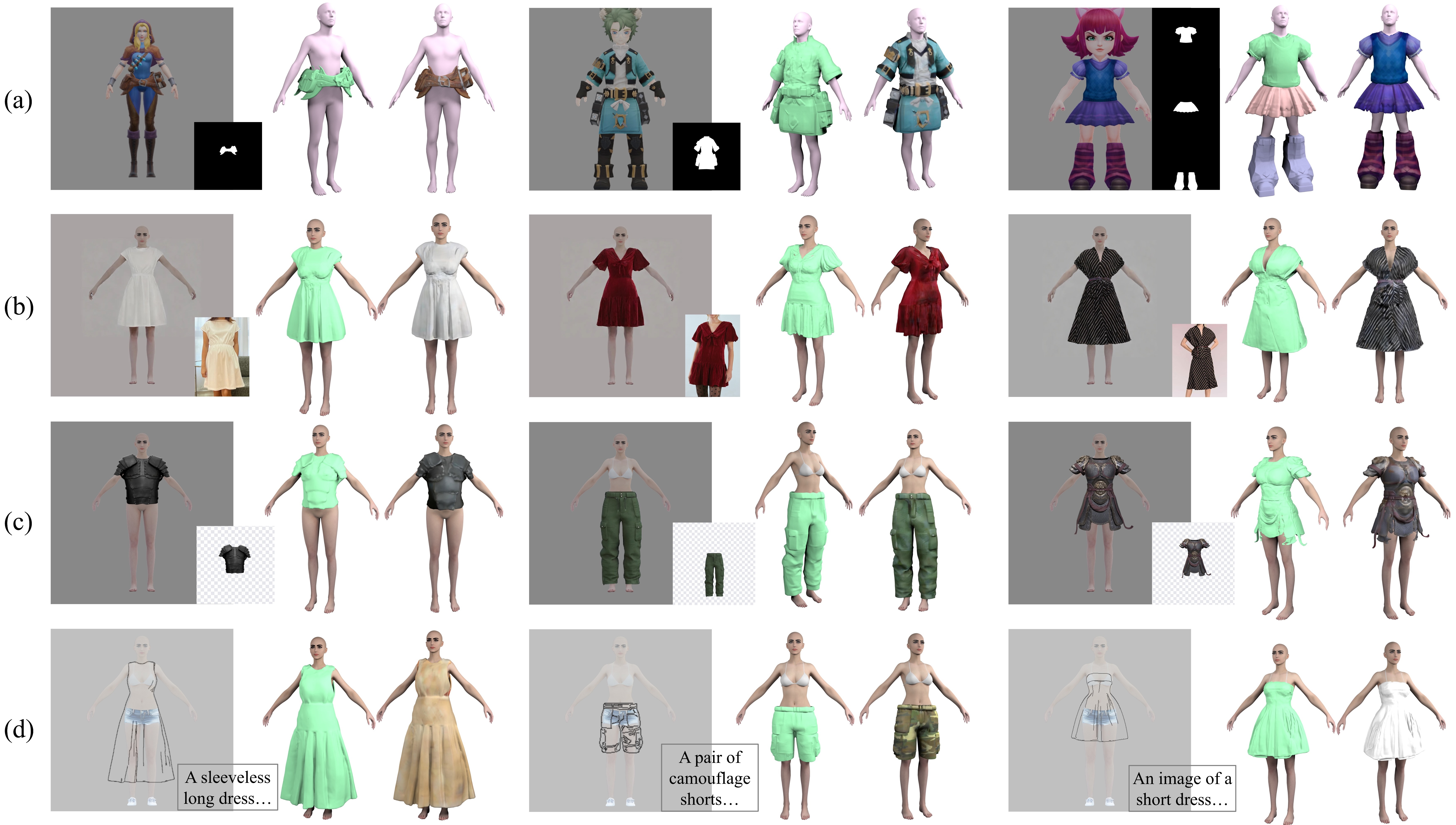}
  \caption{Result gallery across four settings: (a) Single-view reconstruction with fitted SMPLX; (b) Image-Based Virtual Try-On; (c) Assembling existing 2D assets; and (d) Sketch-Based modeling. Each image is followed by the reconstructed asset mesh. Our method demonstrates proficiency in effectively producing body-aligned asset shapes and faithfully capturing high-fidelity geometric details from the input.}
  \label{fig:result_gallery}
\end{figure*}

\medskip
\noindent
\textbf{Result Gallery.} Fig.~\ref{fig:result_gallery} presents our representative results across four settings, which showcases the effectiveness and versatility of our proposed approach in various scenarios.

\section{Limitations and Discussions}\label{sec:conclusion}

This paper presents a method for guiding advanced 3D generative models to directly produce body-aligned 3D assets.
Experimental results indicate that our approach offers significant advantages over existing methods in terms of prompt-following capacity, shape diversity, and shape quality.
We believe that our method opens up a new era in 3D generative model-based garment modeling.
However, a few changes still need to be addressed: 

1) Our method does not explicitly handle multi-layer garment generation. A potential direction for future work is to generate multiple garment layers in parallel and then assemble them on the target human body by resolving asset-to-asset penetrations, as demonstrated in industrial tools such as MetaTailor~\footnote{https://www.metatailor.app/}. We will explore this approach in future research.
2) The mesh topology of the generated asset is based on Marching Cube output, which is not yet production-ready. Since characters are intended to be animatable, topology becomes an even more crucial aspect compared to static objects. Recent researches on artistic style mesh generation, such as \cite{hao2024meshtron}, offers a potential solution for addressing the mesh topology issue.
3) When running SAM, occlusion from the human body or other assets may result in incomplete single-view asset input. While leveraging inpainting tool such as SDXL~\cite{stable_diffusion} can recover from occlusions, the results are not always stable. Future work will focus on developing robust inpainting techniques specifically for the human body asset.
4) Our current approach primarily focuses on main-body fitting. Fine-grained asset generation, such as gloves that involve hand fingers, is not included in this version. This could be accomplished using a fine-grained, hand-aligned 3D asset dataset.

\clearpage

\clearpage

\clearpage

\bibliographystyle{unsrt}
\bibliography{reference}

\end{document}